\documentclass{llncs}
\usepackage[utf8]{inputenc}
\usepackage{hyperref}
\usepackage{booktabs}
\usepackage{wrapfig}
\usepackage{siunitx}

\newif\ifProduction
\newcommand{\ttt}{\texttt}

\newcommand{\clstok}{\ttt{[CLS]}}
\newcommand{\septok}{\ttt{[SEP]}}

\newcommand{\relu}{\mbox{relu}}
\newcommand{\embedq}{\mbox{embed}_{q}}
\newcommand{\embedd}{\mbox{embed}_{d}}
\newcommand{\lnorm}{\mbox{layer-norm}}

\usepackage{color}

\ifProduction
\newcommand{\mynote}[1]{}

\else
\newcommand{\mynote}[1]{\textbf{\color{red}\small #1}}

\fi

\usepackage{xcolor}
\usepackage{soul}

\sethlcolor{black}
\makeatletter
\ifProduction     \else
   \sethlcolor{black}
\fi

\newcommand{\modelone}{Model~1}

\title{Exploring Classic and Neural Lexical Translation Models for Information Retrieval: 
       Interpretability, Effectiveness, and Efficiency  Benefits}
\author{Leonid Boytsov\inst{1}, Zico Kolter\inst{1}\inst{2}}
\institute{Bosch Center for Artificial Intelligence\\\email{leonid.boytsov@us.bosch.com}
           \and
           Carnegie Mellon University\\\email{zkolter@cs.cmu.edu}}
\date{August 2020}

\begin{document}

\maketitle
\thispagestyle{plain}
\begin{abstract}
We study the utility of the lexical translation model (IBM Model 1) for English text retrieval,
in particular, its neural variants that are trained end-to-end.
We use the neural Model1 as an aggregator layer
applied to context-free or contextualized query/document embeddings.
This new approach to design a neural ranking system has benefits for effectiveness,
efficiency, and interpretability.
Specifically, we show that adding an \emph{interpretable} neural Model 1 layer on top of BERT-based contextualized embeddings 
(1) does not decrease accuracy and/or efficiency;
and (2) may overcome the limitation on the maximum sequence length of existing BERT models.
The context-free neural Model 1 is less effective than a BERT-based ranking model,
but it can run efficiently on a CPU 
(without expensive index-time precomputation or query-time operations on large tensors).
Using  Model 1
we produced best neural \emph{and} non-neural runs on the MS MARCO document ranking leaderboard in late 2020.
\end{abstract}

\section{Introduction}
A typical text retrieval system relies on simple term-matching techniques
to generate an initial list of candidates, which can be further re-ranked
using a learned model \cite{buttcher2016information,croft2010search}.
Thus, retrieval performance is adversely affected by a mismatch between query and document terms,
which is known as a vocabulary gap problem~\cite{furnas1987vocabulary,zhao2010term}.
Two decades ago
Berger and Lafferty \cite{berger1999information} proposed to reduce the vocabulary gap
and, thus, to improve retrieval effectiveness with a help of a lexical translation model 
called IBM \modelone\  (henceforth, simply \modelone).
\modelone\  has strong performance when applied to finding answers
in English question-answer~(QA) archives using questions as queries \cite{Jeon2005,RiezlerEtAl2007,surdeanu2011learning,Xue2008}
as well as to cross-lingual retrieval \cite{Zbib2019,lavrenko2002cross}.
Yet, little is known about its effectiveness on \emph{realistic monolingual} English queries,
partly, because training \modelone\  requires large query sets, which previously were not publicly available.

\textbf{Research Question 1.} In the past,  \modelone\  was trained on question-document pairs of similar lengths 
which simplifies the task of finding useful associations between query terms and terms in relevant documents.
It is not clear if \modelone\  can be successfully trained 
if queries are substantially, e.g., two orders of magnitude, shorter than corresponding relevant documents.

\textbf{Research Question 2.}  Furthermore, \modelone\  was trained in a \emph{translation} task
using an expectation-maximization (EM) algorithm \cite{dempster1977maximum,brown1993mathematics} 
that produces a sparse matrix of conditional translation probabilities, 
i.e., a \emph{non-parametric} model.
Can we do better by \emph{parameterizing}
conditional translation probabilities with a neural network  and learning the model \emph{end-to-end} in a \emph{ranking}---rather than a translation---task?

To answer these research questions
we experiment with lexical translation models on two recent MS MARCO collections, 
which have hundreds of thousands of real user queries~\cite{nguyen2016ms,craswell2020overview}.
Specifically, we consider a novel class of ranking models
where an \emph{interpretable} neural \modelone\  layer \emph{aggregates} an output of a token-embedding neural network.
The resulting composite  network (including token embeddings) is learned end-to-end using a ranking objective. 
We consider two scenarios:
 context-independent  token embeddings \cite{collobert2011natural,goldberg2016primer}  and
contextualized token embeddings generated by BERT~\cite{devlin2018bert}.
Note that our approach is \emph{generic}  and can be applied to other embedding networks as well.

The neural \modelone\  layer produces all pairwise similarities $T(q|d)$ for all query and documents BERT word pieces,
which are combined via a straightforward product-of-sum formula without any learned weights:
\begin{equation}
P(Q|D)=\prod\limits_{q \in Q}  \sum\limits_{d \in D} T(q|d) P(d|D), \label{EqLayerIntro}
\end{equation}
where $P(d|D)$ is a maximum-likelihood estimate of the occurrence of $d$ in $D$.
Indeed, a query-document score is a product of scores for individual query word pieces, 
which makes it easy to pinpoint word pieces with largest contributions.
Likewise, for every query word piece
we can easily identify document word pieces with highest contributions to its score.
This makes our model \emph{more interpretable} compared to prior work.

Our contributions can be summarized as follows:
\begin{enumerate}
    \item Adding an \emph{interpretable} neural \modelone\  layer on top of BERT entails virtually 
          no loss in accuracy \emph{and} efficiency
          compared to the vanilla BERT ranker, which is not readily interpretable.
    \item In fact, for long documents the BERT-based \modelone\  may outperform 
           baseline models applied to truncated documents, thus,
           overcoming the limitation on the maximum sequence length 
           of existing pretrained Transformer \cite{vaswani2017attention} models.
           However, evidence was somewhat  inconclusive and we 
           found it was also not conclusive for previously proposed CEDR \cite{CEDR2019} models 
           that too incorporate an aggregator layer (though a \emph{non}-interpretable one); 
    \item A fusion of the non-parametric \modelone\  with BM25 scores can 
          outperform the baseline models, though the gain is modest ($\approx 3$\%).
          In contrast, the fusion with the context-free neural \modelone\  can be substantially ($\approx 10$\%) 
          more effective than the fusion with its non-parametric variant. 
          We show that the neural \modelone\  can be sparsified
          and executed on a CPU more than $10^3$ times
          faster than a BERT-based ranker on a GPU.
          We can, thus, improve the first retrieval stage 
          \emph{without expensive} index-time precomputation approaches.% such as doc2query \cite{nogueira2019document}.
\end{enumerate}

\section{Related Work}

\textit{Translation Models for Text Retrieval.} 
This line of work begins with an influential paper by Berger and Lafferty~\cite{berger1999information}
who first applied \modelone\  to text retrieval~\cite{berger1999information}. 
It was later proved to be useful for finding answers in \emph{monolingual} QA archives
 \cite{Jeon2005,RiezlerEtAl2007,surdeanu2011learning,Xue2008}
 as well as for cross-lingual document retrieval \cite{Zbib2019,lavrenko2002cross}.
% In that, the model is trained on a large \emph{monolingual} collection
% of questions paired with user-provided answers, i.e., on a \emph{parallel} monolingual corpus.
% \modelone\  has been also useful for cross-lingual retrieval \cite{Zbib2019,lavrenko2002cross}.
\modelone\  is a \emph{non-parametric} and \emph{lexical} translation model
that learns \emph{context-independent} translation probabilities of lexemes (or tokens)
from a set of paired documents called a \emph{parallel corpus} or \emph{bitext}.
The learning method is a variant of the expectation-maximization (EM) algorithm~\cite{dempster1977maximum,brown1993mathematics}.

A generic approach to improve performance of non-parametric statistical learning
models consists in parameterizing respective probabilities using neural networks.
An early successful implementation of this idea in language processing
were the hybrid HMM-DNN/RNN systems for speech recognition \cite{bourlard1994connectionist,hinton2012deep}.
More concretely, 
our proposal to use the neural \modelone\   as a last network layer
 %---to compute query-document likelihoods---
was inspired by  the LSTM-CRF  \cite{HuangXY15} and CEDR \cite{CEDR2019} architectures.

There is prior history of applying the neural \modelone\ to retrieval,
however, without training the model on a ranking task.
Zuccon et al.~\cite{zuccon2015integrating} computed translation probabilities
using the cosine similarity between word embeddings (normalized over the sum of similarities for top-$k$ closest words). 
They achieved modest 3-7\% gains on four \emph{small-scale} TREC collections.
Ganguly et al.~\cite{ganguly2015word} used a nearly identical approach  (on similar TREC collections)
and reported slightly better (6-12\%) gains.
Neither Zuccon et al.~\cite{zuccon2015integrating} nor Ganguly et al.~\cite{ganguly2015word} 
attempted to learn translation probabilities from a large set of real user queries.

Zbib et al. \cite{Zbib2019} employed a context-\emph{dependent} lexical neural translation model
for \emph{cross-lingual} retrieval.
They first learn context-dependent translation probabilities
from a bilingual parallel corpus in a lexical \emph{translation} task.
Given a document, highest translation probabilities together with respective tokens
are precomputed in advance and stored in the index.
%ugh we have a better argument against their approach
%This precomputation increases indexing time and size.
Zbib et al. \cite{Zbib2019} trained their model on aligned sentences of similar lengths.
In the case of \emph{monolingual retrieval}, however,
we do not have such fine-grained training data as queries are paired only with much longer relevant documents. 
To our knowledge, there is no \emph{reliable} way to obtain sentence-level relevance labels from this data.

\textit{Neural Ranking}
models have been a popular topic in recent years \cite{guo2019deep},
but the success of early approaches---which predate BERT---was controversial \cite{lin2019neural}.
This changed with adoption of large pretrained models \cite{peters2018deep},
especially after the introduction of the Transformer models \cite{devlin2018bert}  and 
release of BERT \cite{devlin2018bert}.
Nogueira and Cho were first to apply BERT to ranking of text documents \cite{nogueira2019passage}.
In the TREC 2019 deep learning track \cite{craswell2020overview} as well as on the MS MARCO leaderboard \cite{msmarcolb}, BERT-based models outperformed all other approaches by a large margin.

% Before Transformers, 
% LSTMs \cite{HochreiterS97,GersSC00} 
% and other recurrent neural networks
% were a major approach for classification and translation of sequence data.
% They do not have an inherent limitation on a sequence length.
% However, the recently proposed Transformer model \cite{vaswani2017attention} 
The Transformer model \cite{vaswani2017attention} uses an attention mechanism \cite{BahdanauCB14}
where each sequence position can attend to all the positions in the previous layer.
Because self-attention complexity is quadratic with respect to a sequence length,
 Transformer models (BERT including) support only limited-length inputs.
A number of proposals---see Tay et al. \cite{Tay2020} for a survey---aim
to mitigate this constraint, which is complementary to our work. 
 
To process longer documents with existing pretrained models, 
one has to split documents into several chunks, 
process each chunk separately, and aggregate results, e.g., by computing a maximum or a weighted 
prediction score \cite{YilmazWYZL19,DaiC19}.
Such models cannot be trained end-to-end on full documents.
Furthermore, a training procedure has to assume  that each chunk in a relevant document is relevant as well,
which is not quite accurate.
To improve upon simple aggregation approaches, 
MacAvaney et al.~\cite{CEDR2019} 
combined output of several document chunks using
three simpler models: KNRM \cite{XiongDCLP17}, PACRR \cite{HuiYBM18}, and DRMM~\cite{GuoFAC16}.
A more recent PARADE architectures use even simpler aggregation approaches \cite{PARADE2020}.
However, none of the mentioned aggregator models is interpretable
and we propose to replace them with our neural \modelone\  layer.

% There is a number of proposals to mitigate this constraint, e.g.,
% a Conformer-kernel ranking model uses the so-called separable attention \cite{Conformer2020}.
% This line of research  is complementary to our work and 
% we address the reader to a recent survey of such approaches~\cite{Tay2020}.

\textit{Interpretability and Explainability} of statistical models has become a busy area of research.
However, a vast majority of approaches rely on training a separate explanation model
or exploiting saliency/attention maps \cite{Lipton18,rudin2019stop}.
This is problematic, because explanations provided by extraneous models cannot be verified and, thus, trusted \cite{rudin2019stop}.
Moreover, saliency/attention maps reveal which data parts are being processed by a model,
but not \emph{how} the model processes them \cite{SerranoS19,JainW19,rudin2019stop}.
Instead of producing unreliable post hoc explanations, 
Rudin~\cite{rudin2019stop} advocates for networks whose computation is transparent \emph{by design}.
If full transparency is not feasible,
there is still a benefit of \emph{last-layer} interpretability.

In text retrieval we know only two implementations of this idea.
Hofst{\"{a}}tter et al.~\cite{HofstatterZH20} use a kernel-based formula by Xiong et al.~\cite{XiongDCLP17} 
to compute soft-match counts over contextualized embeddings.
Because each pair of query-document tokens produces several soft-match values corresponding to 
different thresholds, it is problematic to aggregate these values in an explainable way.
Though this approach does offer insights into model decisions, 
the aggregation formula is a relatively complicated two-layer neural network with a non-linear (logarithm) activation function after the first layer \cite{HofstatterZH20}.
ColBERT in the re-ranking mode can be seen as an interpretable interaction layer,
however, unlike the neural \modelone\   its use entails a 3\% degradation in accuracy \cite{KhattabZ20}.

\textit{Efficiency.} 
It is possible to speed-up ranking by deferring some computation to index time. 
They can be divided into two groups.
First, it is possible to precompute separate query and document representations,
which can be quickly combined at query-time in a non-linear fashion \cite{KhattabZ20,Gao2020}.
This method entails little to no performance degradation.
Second, one can generate (or enhance) independent query and document representations 
to compare them via the inner-product computation.
Representations---either dense or sparse---were shown to improve the first-stage retrieval albeit at the cost of expensive indexing processing and some loss in effectiveness.
% In particular, Khattab et al~\cite{abs-2007-00814} show that \emph{dense} representations 
% are inferior to the vanilla BERT ranker \cite{nogueira2019document} in a QA task.

In the case of sparse representations, one can rely on Transformer \cite{vaswani2017attention} models
to generate importance weights for document or query terms \cite{DeepCT2019},
augment documents with most likely query terms  \cite{Nogueira2019FromDT,nogueira2019document},
or use a combination of these methods \cite{macavaney2020expansion}.
Due to sparsity of data generated by term expansion and re-weighting models,
it can be stored in a traditional inverted file 
to improve performance of the first retrieval stage.
However, these models are less effective than the vanilla BERT ranker \cite{nogueira2019document} and
they require \emph{costly} index-time processing.

\section{Methods}
\textit{Token Embeddings and Transformers.}
\label{SectionEmbedsBERTandCEDR}
We assume that an input text is split into small chunks of texts called \emph{tokens}.
A token can be a complete English word, a word piece, or a lexeme (a lemma).
The length of a document $d$---denoted as $|d|$---is measured in the number of tokens.
Because neural networks cannot operate directly on text,
a sequence of tokens
$t_1 t_2 \ldots t_n$ is first converted to a sequences 
of $d$-dimensional embedding vectors $w_1 w_2 \ldots w_n$ by an \emph{embedding} network.
Initially, embedding networks were context independent, i.e.,
each token was always mapped to the same vector \cite{goldberg2016primer,collobert2011natural,MikolovSCCD13}.
Peters et al.~\cite{peters2018deep} demonstrated superiority of \emph{contextualized}, i.e.,
context-dependent, embeddings produced a multi-layer bi-directional LSTM \cite{SchusterP97,HochreiterS97,GersSC00}
pretrained on a large corpus in a \emph{self-supervised} manner.
These were later outstripped by large pretrained Transformers \cite{devlin2018bert,radford2018improving}.

In our work we use two types of embeddings: 
vanilla context-free embeddings (see \cite{goldberg2016primer} for an excellent introduction) and 
BERT-based contextualized embeddings \cite{devlin2018bert}.
Due to space constraints, we do not discuss BERT architecture in detail (see \cite{rush2018annotated,devlin2018bert}  instead).
It is crucial, however, to know the following:
\begin{itemize}
    \item Contextualized token embeddings are vectors of the last-layer hidden state;
    \item BERT operates on word pieces \cite{WuSCLNMKCGMKSJL16} rather than complete words;
    \item The vocabulary has close to 30K tokens and includes two special tokens: \clstok\  (an aggregator)
and \septok (a separator);
    \item \clstok\  is always prepended to every token sequence and its embedding is used as a sequence representation for classification and ranking tasks. 
    % This is redundant, as there's an explanation below
    %To this end, we use an extra task-specific prediction head that consumes an output of the \clstok\  token.
\end{itemize}

The ``vanilla'' BERT ranker uses a single fully-connected layer as a prediction head,
which converts the \clstok\  vector into a scalar.
It makes a prediction based on the following  sequence of tokens:  
\clstok\  $q$ \septok\ $d$ \septok,
where $q$ is a query and $d = t_1 t_2 \ldots t_n$ is a document.
Long documents and queries need to be truncated so that the overall number of tokens does not exceed 512.
To overcome this limitation, MacAvaney~et al.~\cite{CEDR2019} proposed an approach that:
\begin{itemize} 
    \item splits longer documents $d$ into $m$ chunks: $d=d_1 d_2 \ldots d_m$; 
    \item generates $m$ token sequences \clstok\  $q$ \septok\ $d_i$ \septok;
    \item processes each sequence with BERT to generate contextualized embeddings for regular tokens
          as well as for \clstok.
\end{itemize}

The outcome of this procedure is $m$ \clstok-vectors $cls_i$ and $n$ contextualized vectors $w_1 w_2 \ldots w_n$:
one for \emph{each} document token $t_i$.
MacAvaney~et al.~\cite{CEDR2019} explore several approaches to combine these contextualized vectors.
First, they extend the vanilla BERT ranker by making prediction
on the average \clstok\  token: ${1 \over m}\sum_{i=1}^{m} cls_i$.
Second, they use contextualized embeddings as a direct replacement of context-free 
embeddings in the following neural architectures: KNRM \cite{XiongDCLP17}, PACRR \cite{HuiYBM18}, and DRMM \cite{GuoFAC16}.
Third, they introduced a CEDR architecture where the \clstok\  embedding is \emph{additionally} incorporated
into KNRM, PACCR, and DRMM in a model-specific way, which further boosts performance.

\textit{Non-parametric \modelone\label{SectionModel1}.}
% Berger and Lafferty \cite{berger1999information} proposed to
% recast retrieval as a translation problem with an objective to learn a soft-matching function 
% that assigns non-zero weights to related but different terms (e.g., synonyms).
Let $P(D|Q)$ denote a probability that a document $D$ is relevant to the query $Q$.
Using the Bayes rule, $P(D|Q)$ is convenient to re-write as $P(D|Q) \propto P(Q|D) P(D)$.
Assuming a uniform prior for the document occurrence probability $p(D)$,
one concludes that the relevance probability is proportional to $P(Q|D)$.
Berger and Lafferty proposed to estimate this probability
with a \emph{term-independent} and \emph{context-free} model
known as \modelone\  \cite{berger1999information}.

Let $T(q|d)$ be a probability that a query token $q$ is a translation of a document token $d$
and $P(d|D)$ is a probability that a token $d$ is ``generated'' by a document $D$.
Then, a probability that query $Q$ is a translation of document $D$ can be computed as 
 a product of individual query term likelihoods as follows:
\begin{equation}\label{EqModel1Simple}
\hspace{-0.5em}\begin{array}{c}
P(Q|D) =\prod\limits_{q \in Q} P(q|D) \\
P(q|D)=\sum\limits_{d \in D} T(q|d) P(d|D)
\\
\end{array}
\end{equation}
The summation in Eq.~\ref{EqClassicModel1} is over \emph{unique} document tokens.
The in-document term probability $P(d|D)$ is a maximum-likelihood estimate.
Making the non-parametric \modelone\  effective requires quite a few tricks.
First,
$P(q|D)$---a likelihood of a query term $q$---is linearly combined with the collection probability $P(q|C)$
using a parameter $\lambda$ \cite{Xue2008,surdeanu2011learning}.
\footnote{$P(q|C)$ is a maximum-likelihood estimate. 
For an out-of-vocabulary term $q$, $P(q|C)$ is set to a small number~(e.g.,~$10^{-9}$).}
\begin{equation}\label{EqClassicModel1}
P(q|D)=(1-\lambda)\left[ \sum_{d \in D} T(q|d) P(d|D)\right] + \lambda P(q|C).
\end{equation}
% Translation probabilities $T(q|d)$ are computed using the EM \cite{brown1993mathematics,dempster1977maximum}
% algorithm implemented in MGIZA \cite{Och2003}.\footnote{\url{https://github.com/moses-smt/mgiza/}}
% MGIZA  models spurious insertions (i.e., a translation from an empty word),
% but we discard them as in prior work \cite{surdeanu2011learning}.
We take several additional measures to improve \modelone\  effectiveness:
\begin{itemize}
    \item We propose to create a parallel corpus by splitting documents and passages
    into small contiguous chunks whose length is comparable to query lengths;
    \item $T(q|d)$ are learned from a symmetrized corpus as proposed  by Jeon et al.~\cite{Jeon2005};
    \item We discard all translation probabilities $T(q|d)$ below an empirically found threshold of about $10^{-3}$ and keep at most $10^6$ most frequent tokens;
    \item We make self-translation probabilities $T(t|t)$ to be equal to an empirically found positive value and rescale  $T(t'|t)$ so that $\sum_{t'} T(t'|t)=1$ as in \cite{Jeon2005,surdeanu2011learning};
\end{itemize}

% We evaluated two splitting methods: 
% \begin{itemize}
%     \item sampling with probabilities proportional to the cosine similarity between query and document terms
% (based on Glove embeddings \cite{PenningtonSM14});
%     \item simple splitting into non-overlapping contiguous sequences containing 
% $\approx \alpha |q|$ tokens, where $\alpha$ is an empirically determined parameter.
% \end{itemize}
% The second, simpler, approach (with $\alpha=1.5$) turned out to be superior.

\textit{Our Neural \modelone.}
% In summary, we use a query-token independent model:
% The score of each query token is computed independently and the scores are multiplied.
% For each token, we first compute a probability-like similarity score between this query token
% and every document token using a neural network. 
% We then compute the query token score using the law of total probability.
Let us rewrite Eq.~\ref{EqModel1Simple}  so that the inner summation is carried out
over all document tokens rather than over the set of unique ones. 
This is particularly relevant for contextualized embeddings where embeddings of identical
tokens  are not guaranteed to be the same (and typically they are not):
\begin{equation}
P(Q|D)=\prod\limits_{q \in Q}  \sum\limits_{i=1}^{|D|} \frac{T(q|d_i)}{|D|}. \label{EqNeuralModelOne}
%P(Q|D)=\prod\limits_{q \in Q}  \sum\limits_{i=1}^{|D|} \frac{T(\embedq{(q)}|\embedd{(d_i)})}{|D|}. \label{EqNeuralModelOne}
%P(Q|D)=\prod\limits_{q \in Q}  \sum\limits_{i=1}^{|D|} \frac{T(\embedq{(q, d_i)}|\embedd{(q, d_i)})}{|D|} \label{EqNeuralModelOne}
\end{equation}
We further propose to compute $T(q|d)$ in Eq.~\ref{EqNeuralModelOne}
by a simple and efficient neural network.
Networks ``consumes'' context-free or contextualized embeddings of tokens $q$ and $d$
and produces a value in the range $[0, 1]$.
To incorporate a self translation probability---crucial for good convergence of the context-free 
model---we set $T(t|t)=p_{self}$ and multiply all other probabilities by  $1-p_{self}$.
However, it was not practical to scale conditional probabilities to ensure that 
 $\forall t_2\; \sum_{t_1} T(t_1|t_2) = 1$.
Thus, $T(t_1|t_2)$ is a similarity function, but not a true probability distribution.
Note that---unlike CEDR \cite{macavaney2020expansion}---we do not use the embedding of the \clstok\  token.

We explored several approaches to neural parametrization of $T(t_1|t_2)$.
Let $\embedq(t_1)$ and $\embedd(t_2)$  denote 
embeddings of query and document tokens, respectively.
One of the simplest approaches is to learn separate embedding networks for queries
and documents and use the scaled cosine similarity:
$$T(t_1|t_2) = 0.5\{\cos(\embedq(t_1), \embedd(t_2)) + 1\}.$$
However, this neural network is not sufficiently expressive and the resulting 
context-free \modelone\   is inferior to the non-parametric \modelone\  learned via EM.
We then found that a key performance ingredient was a concatenation of embeddings with their Hadamard product,
which we think helps the following layers discover better interaction features.
We  pass this combination through one or more fully-connected linear layer with RELUs \cite{Hahnloser98}
followed by a sigmoid:
$$
\begin{array}{l}
T(q|d) = \sigma(F_3(\relu(F_2(\relu(F_1([x_q, x_d, x_q \circ x_d])))))) \\
x_q = P_q(\tanh(\lnorm(\embedq(q))))  \\
x_d = P_d(\tanh(\lnorm(\embedd(d)))),  \\
\end{array}
$$
where $P_q$, $P_d$, and $F_i$ are fully-connected linear  layers; $[x,y]$ is vector concatenation;
 $\lnorm$ is layer normalization \cite{BaKH16}; $x \circ y$ is the Hadamard product.

\label{ExportModel1}
\textit{Neural \modelone\ Sparsification/Export to Non-Parametric Format.}
We can precompute $T(t_1|t_2)$ for all pairs of vocabulary tokens,
discard small values (below a threshold), and store the result as a sparse matrix.
This format permits an extremely efficient execution on CPU (see results in \S \ref{SectionRes}).

\section{Experiments}
\subsection{Setup}
%Learning an effective \modelone\  requires a large set of queries paired with 
%respective relevant documents \anonref. %\cite{boytsov2018efficient}.
% A community QA (CQA) collection \manner\ has a large number of paired question-answer pairs as well.
% Another reason to include it in our experiments,
% because \modelone\   was shown to be effective for CQA data in the past \cite{Jeon2005,RiezlerEtAl2007,surdeanu2011learning,Xue2008}.

\textit{Data sets.} We experiment with MS MARCO collections,
which include data for passage and document retrieval tasks~\cite{nguyen2016ms,craswell2020overview}.
Each MS MARCO collection has a large number of real user queries (see Table \ref{TableDataSets}).
To our knowledge, there are no other collections comparable to MS MARCO in this respect.
% MS MARCO collections are related:
% (1)  queries for the document retrieval task are a subset of queries for the passage retrieval task;
% (2) Each passage belongs to some document;
% (3) A document is considered to be relevant when it contains a relevant passage.
The large set of queries is sampled from the log file of the search engine Bing.
In that, data set creators ensured that all queries can be answered using a short text snippet.
These queries are only sparsely judged (about one relevant passage per query).
Sparse judgments are binary: Relevant documents have grade one and all other documents
have grade zero.

\begin{wraptable}{r}{0.5\textwidth}
\vspace{-2em} % This is required, b/c otherwise wraptable screws things up
\scriptsize
\centering
\begin{tabular}{lccc}\toprule
& \scriptsize  documents \hspace{0.25em} & \scriptsize  \hspace{0.25em} passages & \\\midrule
%& \multicolumn{2}{c}{\scriptsize  general statistics} \\\midrule
\scriptsize \# of documents  & \scriptsize   3.2M & \scriptsize   8.8M \\
\scriptsize avg. \# of doc. lemmas &  \scriptsize   476.7  &  \scriptsize   30.6 \\
\scriptsize avg. \# of query lemmas & \scriptsize   3.2  & \scriptsize   3.5  \\ \midrule
& \multicolumn{2}{c}{\scriptsize \# of queries} \\\midrule
\scriptsize    train/fusion  & \scriptsize   10K  & \scriptsize   20K   \\
\scriptsize    train/modeling  & \scriptsize   357K  & \scriptsize   788.7K  \\
\scriptsize    development  & \scriptsize   2500  & \scriptsize   20K  \\
\scriptsize    test  & \scriptsize   2693  & \scriptsize   3000  \\
\scriptsize    TREC 2019  & \scriptsize   100  & \scriptsize   100  \\
\scriptsize    TREC 2020  & \scriptsize   100  & \scriptsize   100  \\
\bottomrule
\end{tabular}
\caption{MS MARCO data set details\label{TableDataSets}}
\vspace{-2em} % This is required, b/c otherwise wraptable screws things up
\end{wraptable}

In addition to large query sets with sparse judgments, 
we use two evaluation sets from TREC 2019/2020 deep learning tracks  \cite{craswell2020overview}.
These query sets are quite small, but they have been thoroughly judged by NIST assessors
\emph{separately} for a document and a passage retrieval task.
TREC NIST judgements range from zero (not-relevant) to three (perfectly relevant).
% Leaderboard test sets are not publicly available.

We randomly split publicly available training and validation sets into the following subsets:
a small training set to train a linear fusion model (\ttt{train/fusion}),
a large set to train neural models and non-parametric \modelone\  (\ttt{train/modeling}),
a development set (\ttt{development}), and a test set (\ttt{MS MARCO test}) containing at most 3K queries.
Detailed data set statistics is summarized in Table \ref{TableDataSets}.
Note that the training subsets were obtained from the original training set, 
whereas the new development and test sets were obtained from the original development set.
The leaderboard validation set is not publicly available.

We processed collections using Spacy 2.2.3 \cite{spacy2} 
to extract tokens (text words) and lemmas (lexemes) from text.
The frequently occurring words and lemmas were filtered out using Indri's list of stopwords \cite{strohmanindri2005},
which was expanded to include a few contractions such as ``n't'' and ``'ll''.
Lemmas were indexed using Lucene 7.6.
We also generated sub-word tokens, namely BERT word pieces \cite{WuSCLNMKCGMKSJL16,devlin2018bert}, 
using a  HuggingFace Transformers library (version 0.6.2) \cite{Wolf2019HuggingFacesTS}.
We did \emph{not} apply the stopword list to BERT word pieces.

\textit{Basic Setup.}
We experimented on a Linux server equipped with a six-core (12 threads) i7-6800K 3.4 Ghz CPU, 
125 GB of memory, and four GeForce GTX 1080 TI GPUs.
We used the text retrieval framework \ttt{FlexNeuART} \cite{FlexNeuART},
which is implemented in Java.
It employs Lucene 7.6 with a BM25 scorer \cite{Robertson2004} 
to generate an initial list of candidates, which can be further re-ranked
using either traditional or neural re-rankers.
The traditional re-rankers, including the non-parametric \modelone, are implemented in Java as well.
They run in a \emph{multi-threaded mode} (12 threads) and \emph{fully} utilize the CPU.
The neural rankers are implemented using PyTorch 1.4 \cite{paszke2019pytorch} 
and Apache Thrift.\footnote{\url{https://thrift.apache.org/}}
A neural ranker operates as a standalone \emph{single-threaded} server.
Our software is available online \cite{FlexNeuART}.\footnote{\url{https://github.com/oaqa/FlexNeuART}}

Ranking speed is measured
as the overall CPU/GPU \emph{throughput}---rather than latency---per one \emph{thousand} of documents/passages.
Ranking accuracy is measured using the standard utility \ttt{trec\_eval} provided
by TREC organizers.\footnote{\url{https://github.com/usnistgov/trec_eval}}.
Statistical significance is computed using a two-sided t-test with threshold 0.05.

All ranking models are applied to the candidate list generated by a tuned BM25 scorer \cite{Robertson2004}.
BERT-based models re-rank 100 entries with highest BM25 scores:
using a larger pool of candidates hurts both efficiency and accuracy.
All other models,  including the neural context-free \modelone\  re-rank 1000 entries:
Further increasing the number of candidates does not improve accuracy.

\textit{Training Models.} \label{PageTrainModels}
Neural models are trained using a pairwise margin loss.\footnote{We use the loss reduction type \ttt{sum}.}
Training pairs are obtained by combining known relevant documents
with 20 negative examples selected from a set of top-500 candidates returned by Lucene.
In each epoch, we randomly sample one positive and one negative example per query.
BERT-based models first undergo a target-corpus pretraining \cite{RuderH18}
using a masked language modeling and next-sentence prediction objective \cite{devlin2018bert}.
Then, we train them for one epoch in a ranking task.
We use batch size 16 simulated via gradient accumulation.
Context-free \modelone\  is trained from scratch for 32 epochs using batch size 32.
The non-parametric \modelone\  is trained for five epochs with MGIZA \cite{Och2003}.\footnote{\url{https://github.com/moses-smt/mgiza/}}
Further increasing the number of epochs does not substantially improve results.
MGIZA  computes probabilities of spurious insertions (i.e., a translation from an empty word),
but we discard them as in prior work \cite{surdeanu2011learning}.

We use a small weight decay ($10^{-7})$
and a warm-up schedule 
where the learning rate grows linearly from
zero for 10-20\% of the steps until it reaches
the base learning rate \cite{Mosbach2020-kn,Smith17}.
The optimizer is AdamW \cite{loshchilov2017decoupled}.
For BERT-based models we use different base rates for the fully-connected prediction head ($2\cdot10^{-4}$)
and for the main Transformer layers ($2\cdot10^{-5}$).
For the context-free \modelone\  the base rate is $3\cdot10^{-3}$, which is decayed by 0.9 after each epoch.
The learning rate is the same for all parameters.

The trained \emph{neural} \modelone\  is ``exported'' to 
a non-parametric format by precomputing all
pairwise translation probabilities and 
discarding probabilities smaller than $10^{-4}$.
This sparsification/export procedure takes three minutes and 
the exported model is executed using the same Java code as the non-parametric \modelone.
Each neural model and the sparsified \modelone\  is trained and evaluated for five seeds.
To this end, we compute the value for each query and seed and average query-specific values (over five seeds).
All hyper-parameters are tuned on a development set.

\begin{table}[t]
\centering
\begin{tabular}{@{}l|c@{}ccc|c@{}ccc@{}}
\toprule
 & \multicolumn{4}{@{}c@{}|}{documents} & \multicolumn{4}{@{}c@{}}{passages} \\ \midrule
 & \multicolumn{1}{c@{}}{\scriptsize\begin{tabular}{{@{}c@{}}}MS\\MARCO\\test\end{tabular}} & \multicolumn{1}{@{}c@{}}{\scriptsize\begin{tabular}{{@{}c@{}}}TREC \\ 2019\end{tabular}} & \multicolumn{1}{@{}c@{}}{\scriptsize\begin{tabular}{{@{}c@{}}}TREC \\ 2020\end{tabular}} & 
 \multicolumn{1}{@{}c@{}|}{\scriptsize\scriptsize\scriptsize\begin{tabular}{{@{}c@{}}}rank.\\speed\end{tabular}} 
 & \multicolumn{1}{c@{}}{\scriptsize\begin{tabular}{{@{}c@{}}}MS\\MARCO\\test\end{tabular}} & \multicolumn{1}{@{}c@{}}{\scriptsize\begin{tabular}{{@{}c@{}}}TREC\\ 2019\end{tabular}} & \multicolumn{1}{@{}c@{}}{\scriptsize\begin{tabular}{{@{}c@{}}}TREC\\ 2020\end{tabular}} 
 & \multicolumn{1}{@{}c@{}}{\scriptsize\scriptsize\begin{tabular}{{@{}c@{}}}rank.\\speed\end{tabular}} \\ \midrule
 & \tiny MRR 
 & \multicolumn{2}{c}{\tiny NDCG@10 }
 & \multicolumn{1}{c|}{\tiny per 1K} 
 & \tiny MRR 
 & \multicolumn{2}{c}{\tiny NDCG@10 }
 & \multicolumn{1}{c}{\tiny per 1K}  \\ \midrule

 & \multicolumn{8}{c}{\scriptsize \textbf{baselines}} \\ \midrule
\scriptsize BM25 (lemm) & {\scriptsize 0.270\phantom{$^*$}} & {\scriptsize 0.544\phantom{$^*$}} & {\scriptsize 0.524\phantom{$^*$}} & \scriptsize  0.8 \tiny ms & {\scriptsize 0.256\phantom{$^*$}} & {\scriptsize 0.522\phantom{$^*$}} & {\scriptsize 0.516\phantom{$^*$}} & \scriptsize  0.5 \tiny ms \\
\midrule
\scriptsize BM25 (lemm)+BM25 (word) & {\scriptsize 0.274\phantom{$^*$}} & {\scriptsize 0.544\phantom{$^*$}} & {\scriptsize 0.523\phantom{$^*$}} & \scriptsize  2.5 \tiny ms & {\scriptsize 0.265\phantom{$^*$}} & {\scriptsize 0.517\phantom{$^*$}} & {\scriptsize 0.521\phantom{$^*$}} & \scriptsize  0.7 \tiny ms\\
\scriptsize BM25 (lemm)+BM25 (bwps) & {\scriptsize 0.283\phantom{$^*$}} & {\scriptsize 0.528\phantom{$^*$}} & {\scriptsize 0.537\phantom{$^*$}} & \scriptsize  2.2  \tiny ms & {\scriptsize 0.270\phantom{$^*$}} & {\scriptsize 0.518\phantom{$^*$}} & {\scriptsize 0.525\phantom{$^*$}} & \scriptsize  0.9 \tiny ms \\
\midrule
\scriptsize BERT-vanilla (short) & {\scriptsize 0.387\phantom{$^*$}} & {\scriptsize 0.655\phantom{$^*$}} & {\scriptsize 0.623\phantom{$^*$}} & \scriptsize  39 \tiny sec  & \textbf{\scriptsize 0.426\phantom{$^*$}} & {\scriptsize 0.686\phantom{$^*$}} & {\scriptsize 0.684\phantom{$^*$}} & \scriptsize  15 \tiny sec \\
\scriptsize BERT-vanilla (full) & {\scriptsize 0.376$^{\#}$} & {\scriptsize 0.667\phantom{$^*$}} & {\scriptsize 0.631\phantom{$^*$}} & \scriptsize  82 \tiny sec & \scriptsize  & \scriptsize  & \scriptsize  &  \\
\midrule
\scriptsize BERT-CEDR-KRNM & {\scriptsize 0.387\phantom{$^*$}} & {\scriptsize 0.665\phantom{$^*$}} & {\scriptsize 0.649$^{\star}$} & \scriptsize  88 \tiny sec & {\scriptsize 0.421$^{\star}$} & {\scriptsize 0.682\phantom{$^*$}} & {\scriptsize 0.675\phantom{$^*$}} & \scriptsize  16 \tiny ms \\
\scriptsize BERT-CEDR-DRMM & {\scriptsize 0.377$^{\star}$} & {\scriptsize 0.667\phantom{$^*$}} & {\scriptsize 0.636\phantom{$^*$}} & \scriptsize  120 \tiny sec & {\scriptsize 0.425\phantom{$^*$}} & {\scriptsize 0.688\phantom{$^*$}} & \textbf{\scriptsize 0.685\phantom{$^*$}} & \scriptsize  30 \tiny sec \\
\scriptsize BERT-CEDR-PACRR & \textbf{\scriptsize 0.392\phantom{$^*$}} & \textbf{\scriptsize 0.670\phantom{$^*$}} & \textbf{\scriptsize 0.652$^{\star}$} & \scriptsize  81 \tiny sec & {\scriptsize 0.425\phantom{$^*$}} & \textbf{\scriptsize 0.690\phantom{$^*$}} & {\scriptsize 0.684\phantom{$^*$}} & \scriptsize  16 \tiny sec \\
\midrule
 & \multicolumn{8}{c}{\scriptsize \textbf{our methods}} \\ \midrule
\scriptsize BM25 (lemm)+Model1 (word) & {\scriptsize 0.283$^{\star}$} & {\scriptsize 0.548\phantom{$^*$}} & {\scriptsize 0.535\phantom{$^*$}} & \scriptsize  13  \tiny ms& {\scriptsize 0.274$^{\star}$} & {\scriptsize 0.522\phantom{$^*$}} & {\scriptsize 0.567$^{\star}$} & \scriptsize  1.2 \tiny ms\\
\scriptsize BM25 (lemm)+Model1 (bwps) & {\scriptsize 0.284\phantom{$^*$}} & {\scriptsize 0.557\phantom{$^*$}} & {\scriptsize 0.525\phantom{$^*$}} & \scriptsize  33 \tiny ms & {\scriptsize 0.271\phantom{$^*$}} & {\scriptsize 0.517\phantom{$^*$}} & {\scriptsize 0.509\phantom{$^*$}} & \scriptsize  2.7 \tiny ms\\
\midrule
\scriptsize BM25 (lemm)+NN-Model1-exp & {\scriptsize 0.307$^{\star}$} & {\scriptsize 0.568\phantom{$^*$}} & {\scriptsize 0.545\phantom{$^*$}} & \scriptsize  16 \tiny ms & {\scriptsize 0.298$^{\star}$} & {\scriptsize 0.541$^{\star}$} & {\scriptsize 0.581$^{\star}$} & \scriptsize  2.4  \tiny ms \\
\scriptsize BM25 (lemm)+NN-Model1 & {\scriptsize 0.311$^{\star}$} & {\scriptsize 0.566\phantom{$^*$}} & {\scriptsize 0.541\phantom{$^*$}} & \scriptsize  3 \tiny sec & {\scriptsize 0.300$^{\star}$} & {\scriptsize 0.549$^{\star}$} & {\scriptsize 0.587$^{\star}$} & \scriptsize  0.32 \tiny sec \\
\midrule
\scriptsize BERT-Model1 (short) & {\scriptsize 0.384\phantom{$^*$}} & {\scriptsize 0.657\phantom{$^*$}} & {\scriptsize 0.631\phantom{$^*$}} & \scriptsize  36 \tiny sec & \textbf{\scriptsize 0.426\phantom{$^*$}} & {\scriptsize 0.685\phantom{$^*$}} & {\scriptsize 0.682\phantom{$^*$}} & \scriptsize  16 \tiny sec \\
\scriptsize BERT-Model1 (full) & {\scriptsize 0.391$^{\#}$} & {\scriptsize 0.666\phantom{$^*$}} & {\scriptsize 0.637$^{\star}$} & \scriptsize  80 \tiny sec & \scriptsize  & \scriptsize  & \scriptsize  &  \\

\bottomrule
\end{tabular}
\vspace{1em}
\caption{Evaluation results: \ttt{bwps} denotes BERT word pieces, \ttt{lemm} denotes text lemmas,
and \ttt{word} denotes original words. 
\ttt{NN-Model1} and \ttt{NN-Model1-exp} are the context-free neural \modelone\  models: They use only \ttt{bwps}.
\ttt{NN-Model1} runs on GPU whereas \ttt{NN-Model1-exp} runs on CPU.
Ranking speed is throughput and not latency!
Statistical significance is denoted by $^\star$ and $^\#$. Hypotheses are explained in the main text.
\label{TableResults}}
\end{table}

Because context-free \modelone\  rankers are not strong on their own, we evaluate them in a \emph{fusion} mode.
First, \modelone\   is trained on \ttt{train/modeling}.
Then  we linearly combine a  model score with the BM25 score \cite{Robertson2004}.
Optimal weights are computed on a \ttt{train/fusion} subset
using the coordinate ascent algorithm \cite{Metzler2007} from RankLib.\footnote{\url{https://sourceforge.net/p/lemur/wiki/RankLib/}}
To improve effectiveness of this linear fusion, 
we use \modelone\  \emph{log}-scores \emph{normalized} by the number of query words.
In turn, BM25 scores are normalized by the sum of query-term IDF values (see
\cite{Robertson2004} for the description of BM25 and IDF).
As one of the baselines, 
we use a fusion of BM25 scores for different tokenization approaches (basically a multi-field BM25).
Fusion weights are obtained via RankLib on \ttt{train/fusion}.

%\newpage
%Dropout is set to 0.05. Actually, the Transformer layers have a different dropout.
\subsection{Results}\label{SectionRes}
\textit{Model Overview.}
We compare several models (see Table \ref{TableResults}).
First, we use BM25 scores \cite{Robertson2004} computed for the lemmatized text,
henceforth, \ttt{BM25~(lemm)}.
Second, we evaluate several variants of the context-free \modelone.
The non-parametric \modelone\  was trained for both original words and BERT word pieces:
Respective models are denoted as \ttt{Model1 (word)} and \ttt{Model1 (bwps)}.
The neural context-free \modelone---denoted as \ttt{NN-Model1}---was used only with BERT word pieces.
This model was sparsified and exported to a non-parametric format (see \S~\ref{ExportModel1}),
which runs efficiently on a CPU. We denote it as \ttt{NN-Model1-exp}.
Note that context-free \modelone\  rankers are not strong on their own, thus,
we evaluate them in a \emph{fusion} mode
by combining their scores with \ttt{BM25 (lemm)}.

Crucially, all context-free models incorporate exact term-matching signal
via either the self-translation probability or via explicit smoothing with a word collection
probability (see Eq.~\ref{EqClassicModel1}).
Thus, these models should be compared not only with BM25, 
but also with the fusion model incorporating BM25 scores for original words or BERT word pieces.
We denote these baselines as \ttt{BM25 (lemm)+ BM25 (word)} and  \ttt{BM25 (lemm)+ BM25 (bwps)}, respectively.

As we describe in \S~\ref{SectionEmbedsBERTandCEDR}, 
our contextualized \modelone\   applies the neural \modelone\  layer to the contextualized
embeddings produced by BERT. We denote this model as \ttt{BERT-Model1}.
Due to the limitation of existing pretrained Transformer  models,
long documents need to be split into chunks each of which is processed, i.e., contextualized,
separately. 
This is done in \ttt{BERT-Model1 (full)},
 \ttt{BERT-vanilla} {(full)}, and \ttt{BERT-CEDR} \cite{CEDR2019} models.
 These models operate on (mostly) complete documents: For efficiency reasons
 we nevertheless use only the first 1431 tokens (three BERT chunks).
Another approach is to make predictions on much shorter (one BERT chunk) fragments \cite{DaiC19}.
This is done in \ttt{BERT-Model1 (short)}  and \ttt{BERT-vanilla (short)}.
In the passage retrieval task, all passages are short and no truncation or chunking is needed.
Note that we use a \emph{base}, i.e., 
a 12-layer Transformer \cite{vaswani2017attention} model,
since it is more practical then a 24-layer BERT-large 
and performs at par with BERT-large on MS MARCO data \cite{HofstatterZH20}.

We tested  several hypotheses using a two-sided t-test:
\begin{itemize}
    \item \ttt{BM25 (lemm)+ Model1 (word)} is the same as \ttt{BM25 (lemm)+ BM25 (word)};
    \item \ttt{BM25 (lemm)+ Model1 (bwps)} is the same as \ttt{BM25 (lemm)+ BM25 (bwps)};
    \item \ttt{BERT-Model1 (full)} is the same as  \ttt{BERT-vanilla (short)};
    \item For each \ttt{BERT-CEDR} model, we test if it is the same as  \ttt{BERT-vanilla (short)};
    \item \ttt{BERT-vanilla (full)} is the same as  \ttt{BERT-vanilla (short)};
    \item \ttt{BERT-Model1 (full)} is the same as  \ttt{BERT-Model1 (short)};
\end{itemize}
The main purpose of these tests is to assess if special aggregation layers (including the neural \modelone)
can be more accurate compared to models that run on truncated documents.
In Table~\ref{TableResults} statistical significance is indicated by
a special symbol: the last two hypotheses use $^\#$; all other hypotheses use $^\star$.

\textit{Discussion of Results.}
The results are summarized in Table~\ref{TableResults}. 
% First, in nearly all cases  the models outperform \ttt{BM25 (lemm)}.
% However, improvements are more pronounced on the large tests.
% In particular, 
% on \ttt{MS MARCO test} the best BERT models outperform \ttt{BM25 (lemm)}
% by 45\% and 65\% for the document and passage retrieval tasks respectively
% compared to about 25\% and 30\% on \ttt{TREC 2019/2020}.
First note that 
there is less consistency in results on \ttt{TREC 2019/2020} sets 
compared to \ttt{MS MARCO test} sets.
In that, some statistically significant differences (on \ttt{MS MARCO test}) ``disappear'' on \ttt{TREC 2019/2020}.
\ttt{TREC 2019/2020} query sets are quite small and it is more likely (compared to \ttt{MS MARCO test})
to obtain spurious results.
Furthermore,  the fusion model \ttt{BM25 (lemm)+ Model1 (bwps)} is either worse than the baseline model 
\ttt{BM25 (lemm)+ BM25 (bwps)} or the difference is not significant. 
%In some cases, it is worse than \ttt{BM25 (lemm)}.
\ttt{BM25 (lemm)+ Model1 (word)} is mostly better than the respective baseline, but the gain is quite small.
In contrast, the fusion of the neural \modelone\   with BM25 scores for BERT word pieces
is more accurate on all the query sets.
On the \ttt{MS MARCO test} sets it is 15-17\% better than \ttt{BM25 (lemm)}.
These differences are significant on both \ttt{MS MARCO test} sets as well as on \ttt{TREC 2019/2020}
tests sets for the passage retrieval task.
Sparsification of the neural \modelone\   leads only to a small (0.6-1.3\%) loss in accuracy.
In that, the sparsified model---executed on a CPU---is more than $10^3$ times 
faster than BERT-based rankers, which run on a GPU. 
It is $5\times 10^3\times$ faster in the case of passage retrieval.
In contrast, on a GPU,
the fastest neural model KNRM is only 500 times faster than vanilla BERT \cite{HofstatterH19} (also 
for passage retrieval).
For large candidate sets computation of \modelone\   scores can be further sped up (\S 3.1.2.1 \cite{boytsov2018efficient}).
Thus, \ttt{BM25 (lemm)+NN-Model1-exp} can be useful at the candidate generation stage.

We also compared  BERT-based neural \modelone\ 
 with BERT-CEDR and BERT-vanilla models
 on  the \ttt{MS MARCO test} set for the document retrieval task.
 By comparing \ttt{BERT-vanilla (short)}, 
 \ttt{BERT-Model1} \ttt{(short)},
 and \ttt{BERT-Model1 (full)} 
 we can see that the neural \modelone\ 
layer entails virtually no efficiency or accuracy loss.
In fact, \ttt{BERT-Model1 (full)} is  1.8\% and 1\%
better than \ttt{BERT-Model1 (short)} and \ttt{BERT-vanilla (short)}, respectively.
Yet, only the former difference is statistically significant.

Furthermore, the same holds for \ttt{BERT-CEDR-PACRR}, 
which was shown to outperform \ttt{BERT-vanilla} by MacAvaney et al.~\cite{CEDR2019}.
In our experiments it is 1\% better than  \ttt{BERT-vanilla} \ttt{(short)},
but the difference is neither substantial nor statistical significant.
This does not invalidate results of MacAvaney et al.~\cite{CEDR2019}:
They compared \ttt{BERT-CEDR-PACRR} only with \ttt{BERT-vanilla} \ttt{(full)},
which makes predictions on the averaged \clstok\  embeddings.
However, in our experiments, this model is noticeably worse (by 4.2\%) 
than \ttt{BERT-vanilla (short)}  and the difference is statistically significant.
We think that obtaining more conclusive evidence about the effectiveness of aggregation layers 
requires a different data set where relevance is harder to predict from a truncated document.

\textit{Leaderboard Submissions.} 
We combined \ttt{BERT-Model1} with the strong first-stage pipeline,
which uses Lucene to index documents expanded with doc2query \cite{Nogueira2019FromDT,nogueira2019document}
and re-ranks them using a mix of traditional and \ttt{NN-Model1-exp} scores (our exported neural \modelone).
This first-stage pipeline is about as effective as the Conformer-Kernel model \cite{Conformer2020}. 
The combination model achieved the top place on a well-known leaderboard in November and December 2020.
Furthermore, using the non-parametric \modelone, we produced  the best traditional run in December 2020,
which outperformed several neural baselines \cite{boytsov2020traditional}.
\vspace{-0.25em}
\section{Conclusion}
We study a neural \modelone\   
combined with a context-free or contextualized  embedding network
and show that such a combination has benefits to efficiency, effectiveness, and interpretability.
To our knowledge, the context-free neural \modelone\   
is the only neural model  that can be sparsified to  run efficiently on a CPU (up to $5\times 10^3\times$
faster than BERT on a GPU)
without expensive index-time precomputation or query-time operations on large tensors.
We hope that effectiveness of this approach can be further improved,
e.g., by designing a better parametrization of conditional translation probabilities.

% The interpretable neural \modelone\   layer may also be useful for effective ranking of long documents,
% but conclusive experiments may require data sets with different characteristics than MS MARCO.
% Finally, we showed that the non-parametric \modelone\   can be trained via EM even when
% queries and documents have vastly different lengths, 
% but much better results can be obtained
% by training the context-free neural \modelone.
% Using \modelone\ 
% we produced best neural \emph{and} non-neural \cite{boytsov2020traditional} 
% runs on the MS MARCO document ranking leaderboard in late 2020.

%\newpage

%\bibliographystyle{splncs04}
%\bibliography{main}

\end{document}